%% file: acl2022.tex
\pdfoutput=1

\documentclass[11pt]{article}

\usepackage[]{acl2022}

\usepackage{times}
\usepackage{xspace}
\usepackage{latexsym}
\usepackage{graphicx}
\usepackage{amsmath}
\usepackage{bm}

\usepackage{microtype}
\usepackage{colortbl}
\usepackage{makecell}
\usepackage{array}
\newcommand{\colrecall}{$\text{Col}_{R}$\xspace}
\newcommand{\colprec}{$\text{Col}_{P}$\xspace}
\newcommand{\colfscore}{$\text{Col}_{{F}}$\xspace}
\newcommand{\tabrecall}{$\text{Tab}_{{R}}$\xspace}
\newcommand{\tabprec}{$\text{Tab}_{{P}}$\xspace}
\newcommand{\tabfscore}{$\text{Tab}_{{F}}$\xspace}
\newcommand*{\email}[1]{\tt{\small #1}}

\newcommand{\tabincell}[2]{\begin{tabular}{@{}#1@{}}#2\end{tabular}}

\newcommand{\figref}[1]{Figure~\ref{fig:#1}}
\newcommand{\secref}[1]{\S\,\ref{sec:#1}}
\newcommand{\tabref}[1]{Table~\ref{tab:#1}}

\newcommand\daword{CTA}
\newcommand\da{CTA\xspace}
\newcommand\dafull{\textbf{C}ontextualized \textbf{T}able \textbf{A}ugmentation (CTA)\xspace}
\newcommand\TabelPerturbationFull{\textbf{A}dversarial \textbf{T}able \textbf{P}erturbation (\textbf{ATP})\xspace}
\newcommand\tableperturb{ATP\xspace}
\newcommand\advetafull{\textbf{ADVE}rsarial \textbf{T}able perturb\textbf{A}tion (\textbf{ADVETA})}
\usepackage[T1]{fontenc}

\usepackage[utf8]{inputenc}

\usepackage{microtype}
\usepackage{multirow}
\usepackage{multicol}
\usepackage{booktabs}
\usepackage{mathtools}
\usepackage{hyperref}

\title{Towards Robustness of Text-to-SQL Models Against Natural and Realistic Adversarial Table Perturbation}

\author{ Xinyu Pi$^{1*}$, Bing Wang$^{2}$\thanks{~ Equal contributions during the internship at Microsoft Research Asia.}, Yan Gao$^{3}$, Jiaqi Guo$^{4}$, Zhoujun Li$^{2}$, \textbf{Jian-Guang Lou$^{3}$} \\
	$^{1}$    University of Illinois Urbana-Champaign, Urbana, USA,\\
	$^{2}$    State Key Lab of Software Development Environment, Beihang University\\
	$^{3}$    Microsoft Research Asia\\
	$^{4}$    Xi'an Jiaotong University, Xi'an, China\\
	\email{xinyupi2@illinois.edu;} \email{\{bingwang, lizj\}@buaa.edu.cn} \\  \email{jasperguo2013@stu.xjtu.edu.cn;} 
	\email{\{yan.gao, jlou\}@microsoft.com} 
}

\begin{document}
\maketitle
\begin{abstract}

\input{00-abstract}

\end{abstract}

\input{01-introduction}

\input{03-problem_def}

\input{04-benchmark}

\input{05-CTA}

\input{06-experiments}

\input{07-related_work}

\input{08-conclusion}

\input{09-ethical_considerations}

\bibliography{acl2022}
\bibliographystyle{acl_natbib}

\input{10-appendix}

\end{document}

%% file: 00-abstract.tex
The robustness of Text-to-SQL parsers against adversarial perturbations plays a crucial role in delivering highly reliable applications.
Previous studies along this line primarily focused on perturbations in the natural language question side, neglecting the variability of tables.
Motivated by this, we propose the \TabelPerturbationFull as a new attacking paradigm to measure the robustness of Text-to-SQL models. Following this proposition, we curate ADVETA, the first robustness evaluation benchmark featuring natural and realistic ATPs. All tested state-of-the-art models experience dramatic performance drops on ADVETA, revealing models' vulnerability in real-world practices.
To defend against ATP, we build a systematic adversarial training example generation framework tailored for better contextualization of tabular data.
Experiments show that our approach not only brings the best robustness improvement against table-side perturbations but also substantially empowers models against NL-side perturbations.
We release our benchmark and code at: \href{https://github.com/microsoft/ContextualSP}{https://github.com/microsoft/ContextualSP}.

%% file: 01-introduction.tex
\section{Introduction}
\label{sec:intro}

\begin{figure}[t]
    \centering
	\includegraphics[height=8cm]{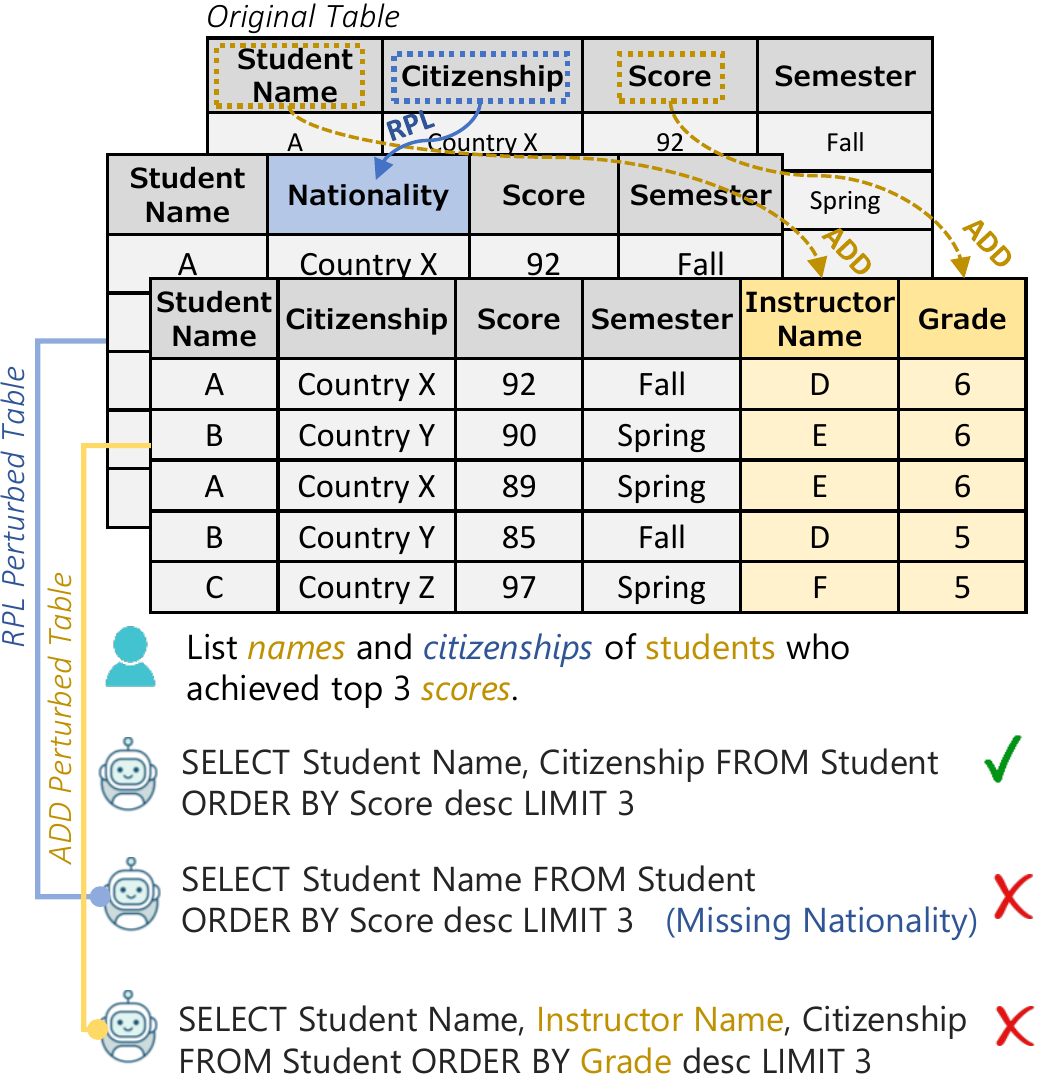}
	\caption{Adversarial examples based on table perturbations for a Text-to-SQL parser. Leaving the NL question unchanged, both replacement of column names (e.g., replace ``Citizenship'' with ``Nationality'') and addition of associated columns (e.g., add ``Instructor Name'' based on ``Student Name''; add ``Grade'' based on ``Score'') mislead the parser to incorrect predictions.}
	\label{fig:demo_tables}
\end{figure}

The goal of Text-to-SQL is to generate an executable SQL query given a natural language (NL) question and corresponding tables as inputs.
By helping non-experts interact with ever-growing databases, this task has many potential applications in real life, thereby receiving considerable interest from both industry and academia~\cite{Li2016UnderstandingNL,zhong2017seq2sql,Affolter2019ACS}.

Recently, existing Text-to-SQL parsers have been found vulnerable to perturbations in NL questions~\cite{gan2021towards, zeng2020photon, deng-etal-2021-structure}.
For example, \citet{deng-etal-2021-structure} removed the explicit mentions of database items in a question while keeping its meaning unchanged, and observed a significant performance drop of a Text-to-SQL parser.
\citet{gan2021towards} also observed a dramatic performance drop when the schema-related tokens in questions are replaced with synonyms.
They investigated both multi-annotations for schema items and adversarial training to improve parsers' robustness against permutations in NL questions.
However, previous works only studied the robustness of parsers from the perspective of NL questions, neglecting variability from the other side of parser input -- tables.

We argue that a reliable parser should also be robust against table-side perturbations since they are inevitably modified in the human-machine interaction process. 
In business scenarios, table maintainers may \textit{(i)} rename columns due to business demands and user preferences.
\textit{(ii)} add new columns into existing tables when business demands change.
Consequently, the extra lexicon diversity introduced by such modifications could harm performances of unrobust Text-to-SQL parsers.
To formalize these scenarios, we propose a new attacking paradigm, \TabelPerturbationFull, to measure parsers' robustness against \textit{natural and realistic} ATPs. In accordance with the two scenarios above, we consider both \textbf{R}E\textbf{PL}ACE (\textbf{RPL}) and \textbf{ADD} perturbations in this work. Figure~\ref{fig:demo_tables} conveys an intuitive understanding of ATP.

Ideally, ATP should be conducted based on two criteria: \textit{(i)} 
Human experts consistently write correct SQL queries before and after table perturbations, yet parsers fail; \textit{(ii)} Perturbed tables look natural and grammatical, and are free from breakage of human language conventions. 
Accordingly, we carefully design principles for RPL/ADD and manually curate the \advetafull~ benchmark based on \textbf{three} existing datasets. 
All evaluated state-of-the-art Text-to-SQL models experience drastic performance drops on ADVETA: On ADVETA-RPL, the average relative percentage drop is as high as $53.1\%$, whereas on ADVETA-ADD is $25.6\%$, revealing models' lack of robustness against ATPs.

Empirically, model robustness can be improved by adversarial training, i.e. re-train models with training set augmented with adversarial examples  ~\cite{Jin2019IsBR}. 
However, due to the different natures of structured tables and unstructured text, well-established text adversarial example generation approaches are not readily applicable. 
Motivated by this, we propose an effective \dafull approach that better leverages tabular context information and carries out ablation analysis.
To summarize, the contributions of our work are three-fold:

\begin{itemize}

\item To the best of our knowledge, we are the \textit{first} to propose definitions and principles of \TabelPerturbationFull as a new attacking paradigm for Text-to-SQL.
    
\item We contribute \textbf{ADVETA}, the first benchmark to evaluate the robustness of Text-to-SQL models. \textit{Significant performance drops} of state-of-the-art models reveals that there is much more to explore beyond high leaderboard scores.
    
\item We design \textbf{CTA}, a systematic adversarial training example generation framework tailored for better contextualization of tabular data. Experiments show that our approach brings model \textit{best robustness gain} and \textit{lowest original performance loss}, compared with various baselines. 
Moreover, we show that adversarial robustness brought by CTA generalizes well to NL-side perturbations.
    
\end{itemize}

%% file: 03-problem_def.tex
\section{Adversarial Table Perturbation}
\label{sec:ctp}

\input{tables/benchmark-statistics}

We propose the \TabelPerturbationFull paradigm to measure robustness of Text-to-SQL models.
For an input table and its associated NL questions, the goal of \tableperturb is to fool Text-to-SQL parsers by perturbing tables naturally and realistically. More specifically, human SQL experts can consistently maintain their correct translations from NL questions to SQL with their understanding of language and table context. 
Formally, \tableperturb consists of two approaches: \textbf{R}E\textbf{PL}ACE (\textbf{RPL}) and \textbf{ADD}. In the rest of this section, we first discuss our consideration of table context, then introduce conduction principles of RPL and ADD.

\subsection{Table Context}

Tables consist of explicit and implicit elements -- both are necessary for understanding table context. ``Explicit elements'' refer to table captions, columns, and cell values. ``Implicit elements'', in our consideration, contains \textbf{T}able \textbf{P}rimary \textbf{E}ntity (\textbf{TPE}) and domain. (Relational) Tables are structured data recording domain-specific attributes (columns) around some central entities (TPE) \cite{sumathi2007fundamentals}. Without the explicit annotation, humans could still make 
correct guesses on them. 
For example, it's intuitive that tables in Figure \ref{fig:demo_tables} can be classified as ``education'' domain, and all of the columns center around the TPE ``student''. Combining both explicit and implicit elements, people achieve an understanding of table context, which becomes the source of lexicon diversity in column descriptions. 

\subsection{REPLACE (RPL) Principles}

Given a target column, the goal of RPL is to seek an alternative column name that makes sense to humans but misleads unrobust models. Gold SQL, as illustrated in Figure \ref{fig:demo_tables}, should be correspondingly adapted by mapping the original column to its new name. In light of this, RPL should fulfill the following two principles:

\textbf{Semantic Equivalency:} Under the table context of target column, substituted column names are expected to convey equivalent semantic meaning as the original name.

\textbf{Phraseology Correctness:} 
ATP aims to be natural and realistic and does not target worst-case attacks.
Therefore, replaced column names are expected to follow linguistic phraseology conventions: \textit{(i)} Grammar Correctness: Substituted column names should be free from grammar errors. \textit{(ii)} Proper Collocation with TPE: New column names should collocate properly with TPE. For example, \textit{height} and \textit{tallness} both collocate well with student (TPE), but conventionally not \textit{altitude}. \textit{(iii)} Idiomaticity: New column names should sound natural to a native speaker to address target columns. For example, \textit{runner-up} means \textit{second place}, and \textit{racer-up} is a bad replacement despite \textit{runner} is synonymous to \textit{racer}. 

\subsection{ADD Principles}
ADD perturbs tables with introductions of new columns. Instead of adding random columns that fit well into the table domain, we pertinently add adversarial columns with respect to a target column for the sake of adversarial efficiency. Gold SQL should remain unchanged after ADD perturbations
\footnote{We omit cell value alignment in ADD for simplicity.}.
Below states ADD principles:

\textbf{Semantic-association \& Domain-relevancy:}
 Given a target column and its table context, newly added columns are expected to \textit{(i)} fit nicely into the table context; \textit{(ii)} have high semantic associations with the target column yet low semantic equivalency (e.g. \textit{sales vs. profits}, \textit{editor vs. author}). 

\textbf{Phraseology Correctness:} Same as RPL, columns should obey human language conventions.

\textbf{Irreplaceability:} 
Unlike RPL, any added columns should be irreplaceable with any original table columns.
In other words, ADD requires semantic equivalency to be filtered out from highly semantic associations.
Otherwise, the original gold SQL will not be the single correct output, which makes the perturbation unreasonable.

%% file: tables/benchmark-statistics.tex
\begin{table*}[t]
\centering
\small
\scalebox{0.85}{
    \begin{tabular}{lccccccccc}
    \toprule
    \multicolumn{1}{c}{\multirow{3}{*}{\bf ADVETA Statistics}} & \multicolumn{3}{c}{\bf Spider} & \multicolumn{3}{c}{\bf WTQ} &
    \multicolumn{3}{c}{\bf WikiSQL}\\
     \cmidrule(lr){2-4}
     \cmidrule(lr){5-7}
     \cmidrule(lr){8-10}
          & \textit{Orig.} & \textit{RPL} & \textit{ADD} & \textit{Orig.} & \textit{RPL} & \textit{ADD} & \textit{Orig.} & \textit{RPL} & \textit{ADD}\\
    \toprule
    \textit{\textbf{~~~~~~~~~~~~~~~~~~Basic Statistics}}\\
    \textit{\#}Total tables & $81$  & $81$  & $81$  &
    $327$ & $327$ & $327$ & $2,716$ & $2,716$ & $2,716$ \\
    \textit{\#}Avg. columns per table & $5.45$  & --  &  -- & $6.31$ & -- & -- & $6.41$ & -- & -- \\
    \textit{\#}Avg. perturbed columns per table & --  & $2.62$  &  $3.64$ & -- & $2.65$ & $3.27$ & -- & $3.70$ & $4.44$ \\
    \textit{\#}Avg. candidates per column & -- & $3.33$  &  $3.97$ & --  & $2.90$ & $3.55$ & -- & $3.32$ & $3.97$ \\
    \textit{\#}Unique columns & $211$  & $911$  &  $1,061$ & $527$ & $1,656$ & $2,976$ & $2,414$ & $10,787$ & $10,474$ \\
    \textit{\#}Unique vocab & $199$ & $598$  & $782$ & $596$ & $1,156$  & $1,459$ & $2,414$ & $4,147$ & $5,099$ \\ 
    \midrule
    \textit{\textbf{~~~~~~~~~~~~~~~Analytical Statistics}}\\
    \textit{\#}Unique semantic meanings & $144$ & $144$  &  $\mathbf{683}$ & $156^*$  & $156^*$ & $\mathbf{702^*}$ & $203^*$ & $203^*$ & $818^*$ \\
    \textit{\#}Avg. col name per semantic meaning & $1.35$ & $\mathbf{6.33}$  & $1.55$ & $1.59^*$  & $\mathbf{5.87^*}$ & $1.64^*$ & $1.67^*$ & $\mathbf{6.12}^*$ & $1.52^*$ \\
    \bottomrule
    \end{tabular}
    }
  \caption{ADVETA statistics comparison between original and RPL/ADD-perturbed dev set. The $^*$ mark denotes that results are based on at most 100 randomly sampled tables and obtained by manual count.}
  \label{tab:bench-stat}%
\end{table*}%

%% file: 04-benchmark.tex
\section{ADVETA Benchmark}
\label{sec:benchmark}

\begin{figure*}[t]
    \centering
    \includegraphics[height=6.5cm, width=16cm]{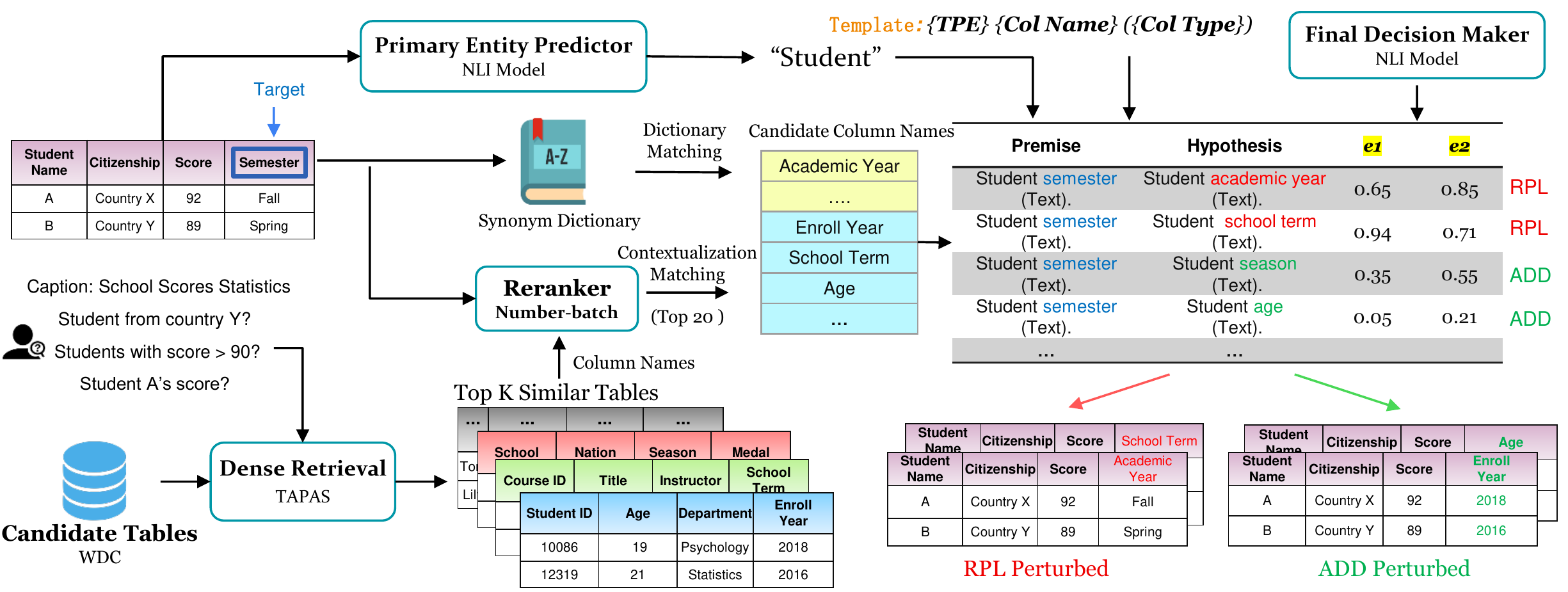}
    \captionsetup{font={small}} 
    \caption{Overview of our \da framework.  In rare cases where TPE is missing, we apply \textit{Primary Entity Predictor} (addressed in \ref{a:tpe_classification}). Otherwise, we simply use annotated TPE. \textit{$e_1$} is obtained with premise-hypothesis as input; \textit{$e_2$} with hypothesis-premise.}
    \label{fig:pipeline}
\end{figure*}

Following RPL and ADD principles, we manually curate the \advetafull~ benchmark based on three mainstream Text-to-SQL datasets, Spider~\cite{yu-etal-2018-spider}, WikiSQL~\cite{zhong2017seq2sql} and WTQ~\cite{papernot2017practical}. 
For each table from the original \textit{development set}, we conduct RPL/ADD annotation separately, perturbing only table columns. 
For its associated NL-SQL pairs, we leave the NL questions unchanged and adapt gold SQLs accordingly.
As a result, ADVETA consists of $3$ (Spider/WTQ/WikiSQL) $*$ $2$ (RPL/ADD) = $6$ subsets.
We next introduce annotation details and characteristics of ADVETA.

\subsection{Annotation Steps}
Five vendors join the annotation process. 
Each base dev set is split into small chunks and is manually annotated by one vendor and reviewed by another, with an inter-annotator agreement to resolve annotation inconsistency.

Before annotation, vendors are first trained to understand table context as described in~\secref{ctp}, then are further instructed of the following details.

\textbf{RPL}: RPL principles are the mandatory requirements. During annotation, vendors are given full Google access to ease the conception of synonymous names for a target column.
\textbf{ADD}: ADD principles will be the primary guideline. Unlike free-style RPL annotations, vendors are provided with a list of 20 candidate columns from where they select 3-5 based on semantic-association\footnote{We generate the candidate list with a retriever-reranker combo from \secref{defense-ccs}.}
Notice that we only consider columns mentioned at least once across NL questions to avoid vain efforts. 
In Appendix \ref{sec:benchmarkexamples}, We display some representative benchmark annotation cases.

\subsection{ADVETA Statistics and Analysis}

We present comprehensive benchmark statistics and analysis results in Table \ref{tab:bench-stat}. Notice that we limit the scope of statistics only to perturbed columns (as marked by \textit{\#Avg. perturbed col per table}). 

\textbf{\textit{Basic Statistics}} reflects elementary information of ADVETA. \textbf{\textit{Analytical Statistics}} illustrate highlighted features of ADVETA compared with original dev-sets: \textit{(i)} Diverse column names for a single semantic meaning: each table from the RPL subset contains approximately five times more lexicons which are used to express a single semantic meaning\footnote{For example, column names \{\textit{Last name, Family name, Surname}\} express a single semantic meaning. In practice, we random sample at most 100 tables from each split, and obtain the number of unique semantic meanings by manual count.}. \textit{(ii)} Table concept richness: each table from ADD subset contains roughly five times more columns with unique semantic meanings.

%% file: 05-CTA.tex
\section{Contextualized Table Augmentation}

\label{sec:defense-ccs}

In this section, we introduce our \dafull framework as an adversarial training example generation approach tailored for tabular data. 
The philosophy of adversarial example generation is straightforward: Pushing augmented RPL/ADD lexicon distributions closer to human-agreeable RPL/ADD distributions.
This requires maximization of lexicon diversity under the constraints of domain relevancy and clear differentiation between semantic association \& semantic equivalency, as stated in ADD principle from \secref{ctp}.

Well-established text adversarial example generation approaches, such as TextFooler \citep{Jin2019IsBR} and BertAttack~\citep{li-etal-2020-bert-attack}, might fail to meet this objective because:
\textit{(i)} They rely on syntactic information (e.g. POS-tag, dependency, semantic role) to perform text transformations.
However, such information is not available in structured tabular data, leading to poor-quality adversarial examples generated by these approaches.
\textit{(ii)} They perform sequential word-by-word  transformations, which could narrow lexicon diversity (e.g.  \textit{written by} will not be replaced by \textit{author}).
\textit{(iii)}
They cannot leverage tabular context to ensure domain relevancy.
\textit{(iv)} They generally fail to distinguish semantic equivalency from high semantic association according to our observations (e.g., fail to distinguish \textit{sales vs. profits}).

To tackle these challenges, we construct the CTA framework.
 Given a \textbf{target column} from a table with NL questions,
 \textit{(i)} a \textbf{dense table retriever} properly contextualizes the input table, thereby pinpointing top-k most \textit{domain-related} tables (and columns) from a large-scale database while \textit{boosting lexicon diversity}.
 \textit{(ii)} A \textbf{reranker} further narrows down \textit{semantic-association} and produces coarse-grained ADD/RPL candidates.
 \textit{(iii)}
 \textbf{NLI decision maker} \textit{distinguishes semantic equivalency from semantic association} and allocates candidate columns to RPL/ADD buckets. 
 A detailed illustration of our CTA framework is shown in \figref{pipeline}. 
 We next introduce each component of CTA.

\subsection{Dense Retrieval for Similar Tables}

The entire framework starts with a dense retrieval module to gather most domain-related tables of user queries. We utilize the Tapas-based~\cite{DBLP:journals/corr/abs-2004-02349} dense retriever in this module~\cite{herzig2021open}, due to its better tabular contextualization expressiveness over classical retrieval methods such as Word2Vec~\cite{mikolov2013distributed} and BM25~\cite{robertson2009probabilistic}. Following the original usage proposed by \citet{DBLP:journals/corr/abs-2004-02349}, we retrieve the top 100 most domain-related tables from the backend Web Data Commons (WDC) \cite{10.1145/2872518.2889386} database consisting of 600k non-repetitive tables with at most $five$ columns.

\subsection{Numberbatch Reranker}

From these retrieved domain-related tables, we further narrow down the range of most semantically associated candidate columns.
This is done by a ConceptNet Numberbatch word embedding \cite{speer2017conceptnet} reranker, who computes the cosine similarity score for a given column pair. We choose ConceptNet Numberbatch due to its advantage of far richer (520k) in-vocabulary multi-grams compared with Word2Vec \cite{mikolov2013distributed}, GloVe \cite{pennington-etal-2014-glove}, and Counter-fitting \cite{mrksic-etal-2016-counter}, which is especially desirable for multi-gram columns. We keep the top 20 similar among them as RPL/ADD candidates for each column of the original table.

\subsection{Word-level Replacement via Dictionary}
Aside from candidates obtained from retriever-reranker for whole-column level RPL, we consider word-level RPL for a target column as a complement. Specifically, we replace each word in a given target column with its synonyms recorded in the Oxford Dictionary (noise is more controllable compared with synonyms gathered by embedding). With a probability $25\%$ for each original word to remain unchanged, we sample until the max pre-defined number (20) of candidates is reached or 5  consecutively repeated candidates are produced.

\subsection{NLI as Final Decision Maker}
\label{sec:nli}
 So far we have pinpointed candidate columns whose domain relevancy and semantic association are already guaranteed. The final stage is to determine which one of RPL/ADD candidates is more suitable for based on its semantic equivalent against target column. Therefore, we leverage RoBERTa-MNLI \cite{DBLP:journals/corr/abs-1907-11692, DBLP:journals/corr/WilliamsNB17}, the expert in \textit{ differentiating semantic equivalency from semantic association}\footnote{We \textbf{\textit{highly recommend}} reading our pilot study in~\ref{a:nli-pilot}.}. Practically, we construct premise-hypothesis by contextualized columns and judge semantic equivalency based on output bidirectional entailment scores $e_1$ and $e_2$.

\paragraph{NLI Premise-Hypothesis Construction}

The Quality of premise-hypothesis plays a key factor for NLI's functioning.
We identify three potentially useful elements for contextualizing columns with surrounding table context: TPE, column type, and column cell value. 
Through manual experiments, we observe that: \textit{(i)} Adding cell value significantly hurt decision accuracy of NLI models. \textit{(ii)} TPE is the most important context information and cannot be ablated. 
\textit{(iii)} Column type information can be a desirable source for word-sense disambiguation. 
Thus the final template for premise-hypothesis construction as python formatted string is expressed as: \textit{$\mathbf{f}``\{\mathbf{TPE}\}~\{\mathbf{CN}\}~(\{\mathbf{CT}\})." $}, where $\mathbf{CN}$ is column name, and $\mathbf{CT}$ is column type.

 \paragraph{RPL/ADD Decision Criterion}

In practice, we observe a discrepancy in output entailment scores between \textit{premise-hypothesis} score $e_1$ and \textit{hypothesis-premise} score $e_2$. Thus we take scores from both direction into consideration. For RPL, we empirically choose $min(e_{1}, e_{2}) >= 0.65$ (\figref{pipeline}) as the final RPL acceptance criterion to reduce occurrences of false positive entailment decision. For ADD, the criterion is instead $max(e_{1}, e_{2}) <= 0.45$ to reduce false negative entailment decisions\footnote{To avoid semantic conflict between a new column $\tilde{c}$ and original columns $c_1,\cdots,c_n$, we apply to each pair of $(\tilde{c}, c_i)$.}.  

%% file: 06-experiments.tex
\section{Experiments and Analysis}

\subsection{Experimental Setup}

\paragraph{Datasets and Models} 
The five original Text-to-SQL datasets involves in our experiments are: Spider \citep{yu-etal-2018-spider}, WikiSQL \citep{zhong2017seq2sql}, WTQ \citep{shi-etal-2020-potential}\footnote{Note that we use the version with SQL annotations provided by \citet{shi-etal-2020-potential} here, since the original WTQ~\citep{pasupat-liang-2015-compositional} only contains answer annotations.}, CoSQL \citep{yu2019cosql} and SParC \citep{yu-etal-2019-sparc}.
Their corresponding perturbed tables are from our ADVETA benchmark. 
WikiSQL and WTQ are single-table, while Spider, CoSQL, and SParC have multi-tables. CoSQL and SParC are known as multi-turn Text-to-SQL datasets, sharing the same tables with Spider.
Dataset statistics are shown in Appendix \tabref{ori-statistics}. 

We evaluate open-source Text-to-SQL models that reach competitive performance on the aforementioned datasets.
DuoRAT~\citep{scholak-etal-2021-duorat} and ETA~\citep{liu-etal-2021-awakening} are baselines for Spider;
SQUALL \citep{shi-etal-2020-potential} is the baseline for WTQ;
SQLova~\citep{hwang2019comprehensive} and CESQL~\citep{guo2019content} are baselines for WikiSQL.
 For the two multi-turn datasets (CoSQL \& SParC), baselines are EditSQL~\citep{zhang-etal-2019-editing} and IGSQL \citep{cai-wan-2020-igsql}. \textit{Exact Match} (\textbf{EM}) is employed for evaluation metric across all settings. Training details are shown in \ref{sec:defense-impl-details}.

\begin{table}[tbp]
  \small
  \centering
    \scalebox{0.85}{
    \begin{tabular}{ p{0.9cm}<{\centering} p{1cm}<{\centering} p{0.5cm}<{\centering} p{2.1cm}<{\centering} p{2.1cm}<{\centering}}
    \toprule
    \textbf{Dataset} & \textbf{Baseline} & \textbf{Dev}  &  \textbf{RPL}   & \textbf{ADD} \\
    \midrule
    \multirow{2}[2]{*}{Spider} & DuoRAT & $69.9$  & $23.8 \pm 2.1$ \color{red}{(-46.1)} & $36.4 \pm 1.3$ \color{red}{(-33.5)} \\
          & ETA   & $70.8$  & $27.6 \pm 1.8$ \color{red}{(-43.2)} & $39.9 \pm 0.9$ \color{red}{(-30.9)} \\
    
    \midrule
    \multirow{2}[2]{*}{WikiSQL} & SQLova & $81.6$  & $27.2 \pm 1.3$ \color{red}{(-54.4)} & $66.2 \pm 2.3$ \color{red}{(-15.4)}\\
        & CESQL & $84.3$  & $52.2 \pm 0.9$ \color{red}{(-32.1)}  & $71.2 \pm 1.5$ \color{red}{(-13.1)} \\
    \midrule 
    WTQ   & SQUALL & $44.1$  & $22.8 \pm 0.5$  \color{red}{(-21.3)} & $32.9 \pm 0.8$ \color{red}{(-11.2)} \\
    
    \midrule
    \multirow{2}[2]{*}{CoSQL} & EditSQL & $39.9$  & $13.3 \pm 0.7$ \color{red}{(-26.6)}  & $30.5 \pm 1.1$ \color{red}{(-9.4)} \\
          & IGSQL & $44.1$  & $16.4 \pm 1.2$ \color{red}{(-27.7)} & $32.8 \pm 2.1$ \color{red}{(-11.3)} \\
    \midrule
    \multirow{2}[2]{*}{SParC} & EditSQL & $47.2$  & $30.5 \pm 0.9$ \color{red}{(-16.7)}  & $40.2 \pm 1.2$ \color{red}{(-7.0)} \\
          & IGSQL & $50.7$  & $34.2 \pm 0.5$ \color{red}{(-16.5)} & $42.9 \pm 1.7$ \color{red}{(-7.8)} \\
    \bottomrule
    \end{tabular}%
    }
  \caption{Results on original dev and ADVETA (RPL and ADD subsets). Red fonts denote \textit{absolute percentage} performance drop compared with original dev.}
  \label{tab:attack-results}%
\end{table}%

\subsection{Attack}

\paragraph{Attack Details}
All baseline models are trained from scratch on corresponding original training sets, and then independently evaluated on original dev sets, ADVETA-RPL and ADVETA-ADD. 
Since columns have around $three$ manual candidates in ADVETA-RPL/ADD, the number of possible perturbed tables scales exponentially with column numbers for a given table from the original dev set.
Therefore, models are evaluated on ADVETA-RPL/ADD by sampling perturbed tables.
For \textbf{\textit{each}} NL-SQL pair and associated table(s), we sample \textbf{\textit{one}} RPL-perturbed table and \textbf{\textit{one}} ADD-perturbed table in each attack.
\textbf{Each} column mentioned from gold SQL is perturbed by a randomly sampled manual candidate from ADVETA. 
For performance stability and statistical significance, we run five attacks with random seeds for each NL-SQL pair.  

\paragraph{Attack Results} 
Table \ref{tab:attack-results} presents the performance of models on original dev sets, ADVETA-RPL and ADVETA-ADD.
Across various task formats, domains, and model designs, state-of-the-art Text-to-SQL parsers experience dramatic performance drop on our benchmark: by RPL perturbations, the relative percentage drop is as high as 53.1\%, whereas on ADD the drop is 25.6\% on average\footnote{Average relative performance presented in Appendix \ref{sec:per-drop-detail}.}. 
Another interesting observation is that RPL consistently leads to higher performance drops than ADD. This is perhaps due to models' heavy reliance on lexical matching, instead of true understanding of language and table context. 
Conclusively, Text-to-SQL models are still far less robust than desired against variability from the table input side.

\begin{table*}[t]
  \centering
  \small
  \scalebox{0.85}{
    \begin{tabular}{
    l 
    p{1.3cm}<{\centering}p{1.3cm}<{\centering}p{1.3cm}<{\centering} c
    p{1.3cm}<{\centering}p{1.3cm}<{\centering}p{1.3cm}<{\centering} c 
    p{1.3cm}<{\centering}p{1.3cm}<{\centering}p{1.3cm}<{\centering}  
    }
    \toprule
    \multirow{2}[1]{*}{Approach} & \multicolumn{3}{c}{WikiSQL} & \phantom{} &  \multicolumn{3}{c}{WTQ} & \phantom{} & \multicolumn{3}{c}{Spider} \\
        \cmidrule{2-4} \cmidrule{6-8} \cmidrule{10-12}
          & \multicolumn{1}{c}{Dev} & \multicolumn{1}{c}{RPL} & \multicolumn{1}{c}{ADD} && \multicolumn{1}{c}{Dev} & \multicolumn{1}{c}{RPL} & \multicolumn{1}{c}{ADD} && \multicolumn{1}{c}{Dev} & \multicolumn{1}{c}{RPL} & \multicolumn{1}{c}{ADD} \\
          
    \toprule
    Orig.   & $81.6$   & $27.2 \pm 1.3$   & $66.2 \pm 2.3$   && $44.1$   & $22.8 \pm 0.5$   & $32.9 \pm 0.8$    && $70.8$   & $27.6 \pm 1.8$  & $39.9 \pm 0.9$  \\
    BA   & $80.1 \pm 0.2$   & $56.8 \pm 0.8$  & $77.9 \pm 0.5$  && $43.9 \pm 0.3$  & $33.6 \pm 0.4$  & $42.8 \pm 0.7$    && $68.1 \pm 0.5$  & $26.9 \pm 1.1$  & $43.1 \pm 0.7$ \\
    TF   & $80.5 \pm 0.3$   & $57.7 \pm 0.7$  & $77.7 \pm 0.4$   && $43.7 \pm 0.4$   & $35.2 \pm 0.5$    & $42.6 \pm 0.6$  && $67.9 \pm 0.6$  & $28.4 \pm 1.2$  & $42.2 \pm 0.6$  \\
    W2V   & $80.8 \pm 0.1$   & $60.7 \pm 1.1$  & $78.2 \pm 0.6$  && $43.4 \pm 0.1$  & $36.8 \pm 0.6$  & $42.2 \pm 0.9$   && $68.3 \pm 0.2$  & $30.1 \pm 1.3$  & $43.3 \pm 1.4$ \\
    MAS   & --   & --   & --   && --   & --   & --   && $69.1 \pm 0.3$  & $27.3 \pm 0.7$  & $35.3 \pm 0.5$  \\
    \midrule
    \daword   & $\bf{81.2 \pm 0.1}$  & $\bf{69.2 \pm 0.5}$   &  $\bf{79.9 \pm 0.3}$  && $\bf{44.1 \pm 0.1}$    & $\bf{41.8 \pm 0.3}$  & $\bf{44.6 \pm 0.5}$  && $\bf{69.8 \pm 0.1}$    & $\bf{35.8 \pm 0.5}$   & $\bf{50.6 \pm 0.1}$ \\
    \quad w/o Retriver   & $81.0 \pm 0.2$  & $68.1 \pm 0.2$  & $78.1 \pm 0.5$  && $44.0 \pm 0.2$  & $40.6 \pm 0.2$  & $42.1 \pm 0.3$  && $69.7 \pm 0.3$  & $34.7 \pm 0.5$   & $43.0 \pm 0.8$  \\
    \quad w/o MNLI   & $80.6 \pm 0.3$  & $61.3 \pm 0.5$   & $78.6 \pm 0.2$   && $43.8 \pm 0.1$   & $36.9 \pm 0.3$    & $43.1 \pm 0.2$ && $69.6 \pm 0.2$   & $29.8 \pm 0.2$  & $47.8 \pm 0.2$ \\
    \bottomrule
    \end{tabular}%
    }
  \caption{Defense results on ADVETA (RPL and ADD subsets). Avg. EM and fluctuations of 5 runs are reported.  Orig. denotes performance without defense from \tabref{attack-results}.}
  
  \label{tab:rpl-add-defense-results}%
\end{table*}%

\begin{table}[tbp]
\small
  \centering
    \scalebox{0.85}{
    \begin{tabular}{ p{2.5cm}<{\centering} p{0.5cm}<{\centering} p{0.9cm}<{\centering} p{0.9cm}<{\centering}}
    \toprule
    \textbf{Schema Linking} & \textbf{Dev} & \textbf{RPL} & \textbf{ADD} \\
    \midrule
     w/o oracle & $70.8$  & $27.6$ \color{red}{(-43.2)}  & $39.9$ \color{red}{(-30.9)} \\
     w/ oracle & $75.2$ & $55.7$ \color{red}{(-19.5)} & $71.3$ \color{red}{(-3.9)} \\
    \bottomrule
    \end{tabular}%
    }
  \caption{Schema linking analysis of ETA on Spider.} 
  \label{tab:oracle-schema-link}%
\end{table}%

\paragraph{Attack Analysis}
To understand the reasons for parsers' vulnerability, we specifically analyze their schema linking modules which are responsible for recognizing table elements mentioned in NL questions.
This module is considered a vital component for Text-to-SQL~\citep{wang-etal-2020-rat, scholak-etal-2021-duorat, liu-etal-2021-awakening}.
We leverage the oracle schema linking annotations on Spider~\citep{lei-etal-2020-examining} and test ETA model on ADVETA using the oracle linkings. 
Note that we update the oracle linkings accordingly when testing on RPL.
Table~\ref{tab:oracle-schema-link} compares the performance of ETA with or without the oracle linkings, from which we make two observations:
\textit{(i)} When guided with the oracle linkings, ETA performs much better on both RPL ($27.6\% \rightarrow 55.7\%$) and ADD ($39.9\% \rightarrow 71.3\%$).
Therefore, the failure in schema linking is one of the essential causes for the vulnerability of Text-to-SQL parsers.
\textit{(ii)} Even with the oracle linkings, the performance of ETA on RPL and ADD still lags behind its performance on the original dev set, especially on RPL.
Through a careful analysis on failure cases, we find that ETA still generates table elements that have a high degree of lexical matching with NL questions, even though the correct table elements are specified in the oracle linkings.

\subsection{Defense}

\paragraph{Defense Details} 
We carry defense experiments with SQLova, SQUALL and ETA on WikiSQL, WTQ and Spider,  respectively. We compare CTA with three baseline adversarial training approaches:
Word2Vec (W2V), TextFooler (TF) \cite{Jin2019IsBR}, and BERT-Attack (BA) \cite{li-etal-2020-bert-attack} (details found in \ref{a:baseline_details}.).
Models are trained from scratch on corresponding \textit{augmented} training sets.
Specifically, for \textbf{\textit{each}} NL-SQL pair, we keep the original table while generating \textbf{\textit{one}} RPL and \textbf{\textit{one}} ADD adversarial example. As a result, augmented training data is three times as large in the sense that each NL-SQL pair is now trained against three tables: original, RPL-perturbed, and ADD-perturbed.
In addition to the adversarial training defense paradigm, we also include the manual version of  Multi-Annotation Selection (MAS) by \citet{gan2021towards} on Spider, using their released data. The rest evaluation process is same as attack.

\paragraph{Defense Results} 
Table \ref{tab:rpl-add-defense-results} presents model performance through various defense approaches. 
We get two observations:
\textit{(i) CTA consistently brings better robustness.} 
Compared with other approaches, \daword-augmented models have the best performance across all ADVETA-RPL/ADD settings, as well as on all original dev sets.
These results demonstrate CTA can effectively improve the robustness of models against RPL and ADD perturbations while introducing fewer noises into original training sets.
Interestingly, we observe that textual adversarial example generation approaches (BA, TF) are outperformed by the simple W2V approach. This verifies our analysis stated in \secref{defense-ccs}.
We include a case study in Appendix \ref{sec:perturbation-case-study} on characteristics of various baselines.

\textit{(ii) CTA fails to bring models back to their original dev performance.} 
Even if trained with high-quality data augmented by \daword, models could still be far worse than their original performance. 
This gap is highly subjected to the similarity of lexicon distribution between train and dev set. 
Concretely, on WikiSQL and WTQ where train and dev set have a similar domain, both RPL performance and ADD performance are brought back closer to original dev performance when augmented with \da. 
On the contrary, on Spider where train-dev domains overlap less, there is still a notable gap between performance after adversarial training and the original dev performance. 
In conclusion, more effective defense paradigms are yet to be investigated.

\begin{table}[tbp]
  \centering
  \small{
  \setlength{\tabcolsep}{4pt}
    \scalebox{0.85}{
    \begin{tabular}{lrrrrrr}
    \toprule
    \textbf{Method} & \multicolumn{1}{l}{\textbf{Col$_{P}$}} & \multicolumn{1}{l}{\textbf{Col$_{R}$}} & \multicolumn{1}{l}{\textbf{Col$_{F}$}} & \multicolumn{1}{l}{\textbf{Tab$_{P}$}} & \multicolumn{1}{l}{\textbf{Tab$_{R}$}} & \multicolumn{1}{l}{\textbf{Tab$_{F}$}} \\
    \midrule
    ETA & $85.4$  & $36.8$  & $51.4$  & $61.3$  & $63.4$  & $62.3$ \\
    W2V$_{RPL}$ & $86.1$  & $40.2$  & $54.8$  & $70.4$  & $72.6$  & $71.5$ \\
    \daword$_{RPL}$ & $\textbf{88.1}$ & $\textbf{50.8}$ & $\textbf{64.4}$ & $\textbf{80.1}$ & $\textbf{85.4}$ & $\textbf{82.7}$ \\
    \midrule
    ETA & $86.3$  & $60.2$  & $70.9$  & $71.2$  & $75.8$  & $73.4$ \\
    W2V$_{ADD}$ & $86.5$  & $63.7$  & $73.4$  & $75.9$  & $82.1$  & $78.9$ \\
    \daword$_{ADD}$ & $\textbf{88.1}$ & $\textbf{70.2}$ & $\textbf{78.2}$ & $\textbf{83.6}$ & $\textbf{89.5}$ & $\textbf{86.4}$ \\
    \bottomrule
    \end{tabular}%
    }
  }
    \caption{The schema linking analysis of attacking with ETA and two defense approaches, namely W2V and \da on Spider; Col as column and Tab as table. P, R, F is short for precision, recall and F1 score, respectively.}
  \label{tab:schema-linking-F1}%
\end{table}%

\paragraph{Defense Analysis} 
Following attack analysis, we conduct schema linking analysis with ETA model augmented with top 2 approaches (i.e. W2V \& CTA) on Spider. 
We follow metric calculation of \citep{liu-etal-2021-awakening} and details are shown in \secref{schema-linking-cal}.
As shown in Table \ref{tab:schema-linking-F1}, both approaches improve the schema linking F$_1$. 
Specifically, CTA improves column F$_1$ by $3\%\sim8\%$, and table F$_1$ by $13\%\sim20\%$, compared with vanilla ETA. This reveals that improvement of robustness can be primarily attributed to better schema linking.

Some might worry about the validity of the CTA's effectiveness due to data leakage risks incurred by the annotation design that vendors are given CTA-retrieved candidate list for ADD annotations. However, we emphasize that: \textit{(i)} RPL have \textit{NO} vulnerability to data leakage since it is entirely independent of CTA. \textit{(ii)} The leakage risk in ADD is negligible. On the one hand, our vast-size (600k tables) backend DB supplies tremendous data diversity, maximally reducing multiple retrievals of a single table; On the other hand, CTA's superior performance on Spider, the representative feature of which is cross-domain \& cross-database across train-test splits (thus makes performance gain from data leakage hardly possible), further testifies its authentic effectiveness.

\subsection{CTA Ablation Study}
We carry out an ablation study to understand the roles of two core components of \daword: dense retriever and RoBERTa-MNLI. Results are shown in \tabref{rpl-add-defense-results}. 

\paragraph{\da w/o Retriever} 
RPL candidates are generated merely from the dictionary; ADD generation is the same as W2V baseline. Compared with complete CTA, models augmented with this setting experience $1.1\%\sim1.2\%$ and $1.8\%\sim7.6\%$ performance drop on RPL and ADD, respectively. We attribute RPL drops to loss of real-world lexicon diversity and ADD drops to loss of domain relevancy.

\paragraph{\da w/o MNLI} RPL and ADD candidates are generated in the same way as \daword, but without denoising of MNLI. 
RPL/ADD decisions solely rely on ranked semantic similarity. 
Compared with complete CTA, models augmented by this setting experience significant performance drops ($4.9\%\sim7.9\%$) on all RPL subsets, and moderate drops ($1.5\%\sim2.8\%$) on all ADD subsets.
We attribute these drops to the inaccurate differentiation between semantic equivalency and semantic association due to lack of MNLI, which results in noisy RPL/ADD adversarial examples.

\subsection{Generalization to NL Perturbations}

\begin{table}[tbp]
  \centering
  \small
    \scalebox{0.9}{
        \begin{tabular}{lcc}
        \toprule
        Model & \multicolumn{1}{l}{Spider} & \multicolumn{1}{l}{Spider-Syn} \\
        \midrule
        $\textnormal{RAT-SQL}_{\textnormal{\tiny{BERT}}}$ \cite{wang-etal-2020-rat} & 69.7  & 48.2 \\
        $\textnormal{RAT-SQL}_{\textnormal{\tiny{BERT}}}$+MAS \cite{gan2021towards} & 67.4  & 62.6 \\
        \midrule
        \textsc{ETA} \cite{liu-etal-2021-awakening}   & 70.8  & 50.6 \\
        \textsc{ETA+\daword} & 69.8  & 60.4 \\
        \bottomrule
        \end{tabular}%
    }
  \caption{EM on Spider/Spider-Syn dev-sets.}
  \label{tab:spider-syn}%
\end{table}%

Beyond CTA's effectiveness against table-side perturbations, a natural question follows: could re-training with adversarial table examples improve model robustness against perturbations from the other side of Text-to-SQL input (i.e., NL questions)?  
To explore this, we directly evaluate ETA (trained with CTA-augmented Spider train-set) on Spider-Syn dataset~\citep{gan2021towards}, which replaces schema-related tokens in NL question with its synonym.
We observe an encouraging $9.8\%$ EM improvement compared with vanilla ETA (trained with Spider train-set). This verifies \textit{CTA's generalizability to NL-side perturbations}, with comparable effectiveness as the previous SOTA defense approach MAS, which fails to generalize to table-side perturbations on ADVETA in \tabref{rpl-add-defense-results}.

%% file: 07-related_work.tex
\section{Related Work}
\paragraph{Robustness of Text-to-SQL}

As discussed in \secref{intro}, previous works \cite{gan2021towards, zeng2020photon, deng-etal-2021-structure} exclusively study robustness of Text-to-SQL parsers against perturbations in NL question inputs. 
Our work instead focuses on variability from the table input side and reveals parsers' vulnerability to table perturbations.

\paragraph{Adversarial Example Generation}

Existing works on adversarial text example generations can be classified into three categories:
(1) Sentence-Level. This line of work generates adversarial examples by introducing distracting sentences or paraphrasing sentences~\citep{jia-liang-2017-adversarial,iyyer-etal-2018-adversarial}.
(2) Word-Level. This dimension of work generates adversarial examples by flipping words in a sentence, replacing words with their synonyms, and deleting random words~\citep{li-etal-2020-bert-attack,ren-etal-2019-generating,Jin2019IsBR}.
(3) Char-Level. This line of work flips, deletes, and inserts random chars in a word to generate adversarial examples~\cite{Belinkov2018SyntheticAN,Gao2018BlackBoxGO}.
All the three categories of approaches have been widely used to reveal vulnerabilities of high-performance neural models on various tasks, including text classification \cite{ren-etal-2019-generating, morris-etal-2020-textattack}, natural language inference \cite{li-etal-2020-bert-attack} and question answering \cite{ribeiro-etal-2018-semantically}.
Previous work on robustness of Text-to-SQL and semantic parsing models primarily adopt word-level perturbations to generate adversarial examples~\citep{huang-etal-2021-robustness}.
For example, the Spider-Sync adversarial benchmark~\citep{gan2021towards} is curated by replacing schema-related words in questions with their synonyms.

Despite these methods' effectiveness in generating adversarial text examples, they are not readily applicable for structural tabular data, as we discussed in \secref{defense-ccs}.
Apart from this, table-side perturbations enjoy \textbf{\textit{much higher attacking efficiency}}: the attack coverage of a single table modification includes all affiliated SQLs, whereas one NL-side perturbation only affects a single SQL. Combined with the lighter cognitive efforts of tabular context understanding than NL-understanding, ATP is \textbf{\textit{arguably lower in annotation costs}}.  

Previous work on table perturbations~\citep{DBLP:conf/aaai/CartellaAFYAE21, DBLP:journals/corr/abs-1911-03274} focuses on table cell values; another work, \cite{ma2020mtteql} study impacts of naively (i.e., without consideration of table context information and without human guarantee) renaming irrelevant columns and adding irrelevant columns.
In this work, we focus on table columns and propose an effective CTA framework that better leverages tabular context information for adversarial example generation, as well as manually annotate ADVETA benchmark.

%% file: 08-conclusion.tex
\section{Conclusion}

We introduce \TabelPerturbationFull, a new paradigm for evaluating model robustness on Text-to-SQL and define its conduction principles. 
We curate the ADVETA benchmark, on which all state-of-the-art models experience dramatic performance drop. 
For defense purposes, we design the CTA framework tailored for tabular adversarial training example generation. 
While CTA outperforms all baseline methods in robustness enhancement, there is still an unfilled gap from the original performance. 
This calls for future exploration of the robustness of Text-to-SQL parsers against ATP. 

%% file: 09-ethical_considerations.tex
\section*{Ethical Considerations}
Our ADVETA benchmark presented in this work is a free and open resource for the community to study the robustness of Text-to-SQL models.
We collected tables from three mainstream Text-to-SQL datasets, Spider~\cite{yu-etal-2018-spider}, WikiSQL~\cite{zhong2017seq2sql} and WTQ~\cite{papernot2017practical}, which are also free and open datasets for research use.
For the table perturbation step, we hire professional annotators to find suitable RPL/ADD candidates for target columns.
We pay the annotators at a price of 10 dollars per hour.
The total time cost for annotating our benchmark is 253 hours.

All the experiments in this paper can be run on a single Tesla V100 GPU.
Our benchmark will be released along with the paper.

%% file: 10-appendix.tex
\appendix
\label{sec:appendix}

\section{Benchmark Examples}
\label{sec:benchmarkexamples}

\begin{figure*}[t]
    \centering
	\includegraphics[height=7cm, width=13.5cm]{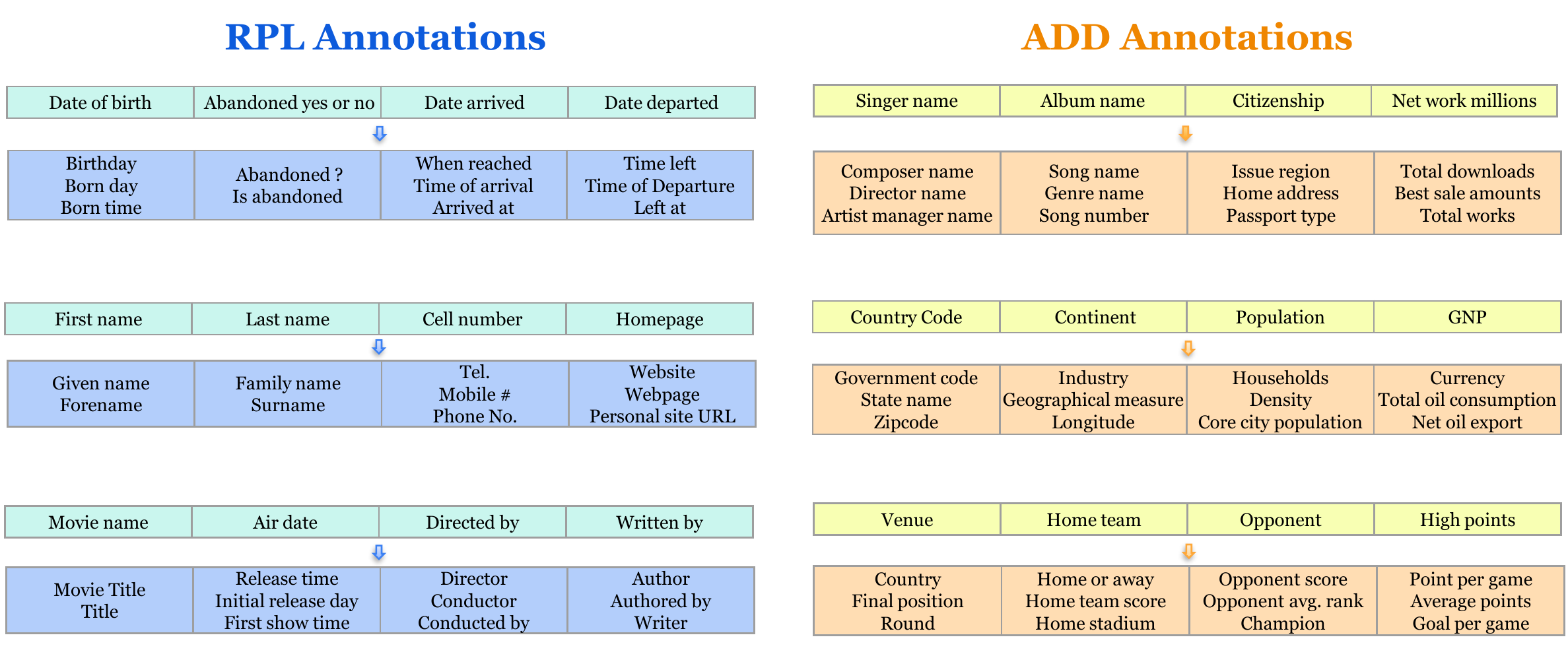}
	\caption{RPL and ADD annotation examples from our ATP benchmark. Rows with shallow colors are original headers, whereas those deep-shaded ones are our human annotations.}
	\label{fig:bm_example}
\end{figure*}

We display some representative benchmark annotation cases for to convey readers a intuitive feeling on our RPL and ADD subsets. As reflected in \figref{bm_example}, RPL reflects the following characteristics beyond RPL principles: \textit{(i)} Abbreviation of common words. e.g. \textit{Cell number} vs. \textit{Tel.} \textit{(ii)} Idiomatic transformation e.g. \textit{Air date} vs. \textit{Release time} \textit{(iii)} Part of speech structure transformation e.g. \textit{Written by} vs. \textit{Author}. ADD perturbations faithfully obey ADD principles and additions demonstrate high coherency with respect to original domain.

\section{CTA Details}

\subsection{NLI-based Substitutability Verification}
\label{a:nli-pilot}
\begin{table}[htbp]
  \centering
    \begin{tabular}{lcccc}
    \toprule
    \textbf{Approach} & $e_1$ & $e_2$ & \textbf{$\Delta_{e_1}$} & \textbf{$\Delta_{e_2}$} \\
    \midrule
    \textbf{\small{ \textit{Roberta-RTE}}} \\
    human & $\mathbf48.5$  & $48.1$  & $0.65$ & $0.46$ \\
    embedding & $45.7$ & $45.6$ & $0.26$ & $0.30$ \\
    ranodm & $43.0$ & $42.8$ & $0.53$ & $0.70$ \\
    \midrule
    \textbf{\small{ \textit{Roberta-SNLI}}} \\
    human & $74.5$  & $74.1$  & $0.48$ & $0.61$ \\
    embedding & $56.7$ & $66.0$ & $0.75$ & $0.37$ \\
    ranodm & $31.2$ & $30.9$ & $0.78$ & $0.64$ \\
    \midrule
    \textbf{\small{ \textit{Roberta-MNLI}}} \\
    human & $\mathbf{77.1}$  & $76.4$  & $0.86$ & $0.36$ \\
    embedding & $\mathbf{52.2}$ & $58.7$ & $0.34$ & $0.69$ \\
    ranodm & $\mathbf{16.5}$ & $14.8$ & $0.50$ & $0.49$ \\

    \bottomrule
    \end{tabular}%
    \caption{Average foward entailment score $e_1$, backward entail $e_2$, and corresponding standard deviations across 9 settings. In all human annotation cases, higher entailment is better. In all random replacement cases, lower is better. }%
  \label{tab:nli}%
\end{table}%

\paragraph{Implementation Details}

For each pair of target column and candidate column, we contextualize each column with the template described in Premise-Hypothesis Construction from section \secref{defense-ccs}. Then with the contextualized target column as the premise and the contextualized RPL candidate as the hypothesis, the NLI model computes both forward entailment score $e1$ and backward score $e2$. Notice that $e2$ computation takes the contextualized  RPL candidate as premise and the contextualized target column as hypothesis in input. We obtain entailment scores from both directions because of the observed score fluctuation caused by reversion in practicable cases. 

\paragraph{Pilot Study for Model Ability}

We carry out a pilot study to test NLI models' capability of differentiating semantic equivalency and similarity in this section. RoBERTa \cite{DBLP:journals/corr/abs-1907-11692} is chosen as the backbone model due to its outstanding performance and computational efficiency across various NLI datasets. Fine-tuned RoBERTa on three well-known NLI datasets: RTE \cite{series/synthesis/2013Dagan}, SNLI \cite{DBLP:journals/corr/BowmanAPM15}, and MNLI \cite{DBLP:journals/corr/WilliamsNB17} are compared to demonstrate model ability difference due to training data,.

 We consider three levels of substitutability, from highest to lowest: human manual substitution (human-annotated replacements sampled from benchmark RPL subsets), embedding-based substitution (top-10 similar multi-grams from ConceptNet Numberbatch word embedding \cite{speer2017conceptnet}), and random substitution (randomly sampled columns across benchmark\cite{speer2017conceptnet}). Practically, we randomly sample 1000 pairs of data each time and repeat each setting five times. 

We report the both average forward $e_1$ and backward entailment scores $e_2$, as well their standard deviations for each setting across five runs (table \ref{tab:nli}). It is immediately obvious that RoBERTa-MNLI surpasses other models in verbal dexterity: the entailment score correlates best with true degrees of substitutability.

\paragraph{Performance on SimLex-999}

\begin{table}[htbp]
  \centering
  \small
    \begin{tabular}{lc}
    \toprule
    \textbf{Approach} & $\rho$ \\
    \midrule
    Word2Vec \citep{mikolov2013distributed}  & $0.37$   \\ 
    Glove \citep{pennington-etal-2014-glove}  & $0.41$ \\
    Glove + Counter-fitting \citep{DBLP:journals/corr/MrksicSTGRSVWY16} & $0.58$\\
    NMT Emedding \citep{hill2015embedding} & 0.58 \\
    aragram-SL999 \citep{wieting-etal-2015-paraphrase} & $0.69$ \\
    RoBERTa-MNLI (ours) & $\mathbf{0.70}$ \\
    \bottomrule
    \end{tabular}%
    \caption{Results on SimLex-999. $\rho$ ( Perason correlation) is used as the primary metric.}%
  \label{tab:nli-perason}%
\end{table}%

\textit{SimLex-999} \cite{hill-etal-2015-simlex} is a gold standard resource for measuring how well models capture similarity, rather than relatedness or association between a input pair of words (e.g. \textit{cold} and \textit{hot} are closely associated but definitely not similar). Thus it is especially suitable for our purpose of further testing RoBERTa-MNLI's ability of semantic discrimination. We treat the entailment score produced by the model as its judgment of semantic similarity and report its Pearson correlation against the ground truth similarity score. Results suggest that RoBERTa-MNLI is quite competitive at discriminating association and relatedness from similarity.

\paragraph{Case Study}
To test the hard case performance of RoBERTa-MNLI, we come up with some tricky examples as shown in \tabref{nli-hard}. The upper half of the table presents hard \textit{\textbf{replaceable}} cases that emphasize idiomatic transformations or word-sense disambiguation. The lower half contains hard \textit{\textbf{irreplaceable}} cases in which phrases have a high degree of conceptual association, yet still not semantically equivalent. Results reveal the surprisingly abundant and accurate lexicon knowledge condensed in RoBERTa-MNLI.

\begin{table}[h]
\centering
\footnotesize
\scalebox{0.90}{
\begin{tabular}{lcccc}
\toprule
Premise & Hypothesis & ENT & NON-ENT \\
\toprule
\textbf{\textit{~Replaceable}}\\
Runner-up. & Second place.  & $\mathbf{97.1}$  & $2.9$ \\
First name. & Given name. & $\mathbf{93.7}$ & $6.3$\\
Airline code. & Airline number. & $\mathbf{82.3}$ & $17.7$\\
Cartoon air date. & Cartoon release time. & $\mathbf{91.4}$ & $8.6$ \\
Book author. & Book written by. & $\mathbf{97.8}$ & $2.2$ \\
\midrule
\textbf{\textit{~Irreplaceable}}\\
Student height. & Student altitude.  & $26.9$  &  $\mathbf{73.1}$\\
Company sales. & Company profits. & $1.9$ & $\mathbf{98.1}$ \\
People killed. & People injured.  & $2.1$  & $\mathbf{97.9}$  \\
Population number. & Population code. & $ 37.1$ & $\mathbf{62.9}$  \\
Political party. & Political celebration. & $ 27.5$ & $\mathbf{72.5}$ \\
\bottomrule
\end{tabular}}
\caption{Hard cases we come up with to explore upper-bounds of Roberta-MNLI ability. ENT as entaiment score, NON-ENT as contradiction + neutral score. Score of expected label is bolded.}
\label{tab:nli-hard}
\end{table}

\subsection{Zero-shot TPE Classification}
We build the previous premise-hypothesis construction in \secref{nli} based on the assumption of availability of TPE, which is frequently not true. Thus our goal is to make a reasonable prediction on TPE for those missing cases. Practically, we make use HuggingFace \citep{wolf-etal-2020-transformers} implementation of zero-shot text classification \citep{DBLP:journals/corr/abs-1909-00161}
to classify missing TPE into 48 pre-defined categories with the input of concatenated table caption, columns, and cell values.

\label{a:tpe_classification}
 \paragraph{Implementation Details} Based on the 60+ fine-grained categories defined in Few-NERD \citep{DBLP:journals/corr/abs-2105-07464}, We modify and integrate them into 48 classes as candidate labels ($|L| = 48$). With a Roberta-MNLI as the workhorse model, our overall modeling process is modeled as

$$
\begin{aligned}
\tilde{c_t} &= \underset{i}{\operatorname{\arg\max}}~~
 \frac{\exp(f_{\theta}(\mathbf{L}_i ~ \vert ~  d; \mathbf{c}; \mathbf{v}; d)_{ent})}{\sum_{j \in |L|} \exp (f_{\theta}(\mathbf{L}_j ~ \vert ~ d; \mathbf{c}; \mathbf{v})_{ent})} 
\end{aligned}
$$
where $\mathbf{c}$ is column names, $\mathbf{v}$ is a randomly selected column value affiliated with a given column, and $d$ is table captions for a given table. Roberta-MNLI (annotated as $f_\theta$) outputs raw logits of contradiction, neutral, and entailment scores. Softmax is finally applied entailment logits across 48 categories, with the top 1 label as final the primary entity prediction.

\paragraph{Human evaluation} We randomly sample 100 tables from our benchmark and ask three vendors to rate the reasonability of each predicted TPE from a scale of $1 - 5$. 1 as totally unreasonable, 3 as mildly acceptable, and 5 as perfectly parallel with human guesses. We average out the rating from all three vendors and get a result of \textit{\textcolor{purple}{4.13}}. This indicates the practicability of zero-shot TPE classification.

\begin{table*}[t]
    \centering
    \small
    \begin{tabular}{llllll}
    \toprule
    Perturbation & Table Context & BA    & TF    & W2V   & \daword \\
    \midrule
    \multirow{3}[6]{*}{RPL} & \tabincell{l}{club id \\ \textcolor{red}{region} \\ name} & \tabincell{l}{member \\ regional \\ district} & \tabincell{l}{districts \\ zones \\ sphere} & \tabincell{l}{regionary \\ location \\ regions} & \tabincell{l}{place \\ location \\ district} \\
\cmidrule{2-6}  & \tabincell{l}{author id \\ \textcolor{red}{type} \\ title} & \tabincell{l}{types \\ number \\ style} & \tabincell{l}{guy \\ genus \\ categories} & \tabincell{l}{typeful \\ example \\ sort} & \tabincell{l}{category \\ genre \\ kind} \\
\cmidrule{2-6}          & \tabincell{l}{singer id \\ \textcolor{red}{song name} \\ country} & \tabincell{l}{songs title \\ singer name \\ chorus name} & \tabincell{l}{ballads denomination \\ ballads appointments \\ song designation} & \tabincell{l}{name \\ polynymous \\ folk-song name} & \tabincell{l}{music name \\ song title \\ music designation} \\
    \midrule
    \multirow{3}[6]{*}{ADD} & \tabincell{l}{course id \\ \textcolor{red}{semester} \\ section id} & \tabincell{l}{classes \\ honors \\ session} & \tabincell{l}{sophomore \\ majoring \\ freshman} & \tabincell{l}{studential \\ intersession \\ undergraduate} & \tabincell{l}{school \\ enrollment \\ university} \\
\cmidrule{2-6}          & \tabincell{l}{artist id \\ \textcolor{red}{artist} \\ age} & \tabincell{l}{composition \\ creator \\ design} & \tabincell{l}{musicianship \\ thespian \\ arranger} & \tabincell{l}{tachiste \\ neosurrealist \\ creative person} & \tabincell{l}{publisher \\ album \\ genre} \\
\cmidrule{2-6}          & \tabincell{l}{movie id \\ \textcolor{red}{director} \\ year} & \tabincell{l}{designer \\ operator \\ composer} & \tabincell{l}{officers \\ padrone \\ guide} & \tabincell{l}{corporate leader \\ supermanager \\ executive} & \tabincell{l}{producer \\ scenarist \\ writer} \\
    \bottomrule
    \end{tabular}%
\caption{Adversarial training examples generated by CTA and baseline approaches. Words with red color font are target columns.}
  \label{tab:case-study}%
\end{table*}%

\subsection{Perturbation Case Study}
\label{sec:perturbation-case-study}

In this section, we present a case study on adversarial training examples generated by CTA and baseline approaches in \tabref{case-study}. We can make the following observations: \textit{(i)} CTA tend to produce less low-frequency words (e.g. \textit{padrone, neosurrealist}) in both RPL and ADD i.e. lower perplexity. \textit{(ii)} CTA-generated samples fit better with the specificity level of table columns. For example, RPL pair \textit{(region, sphere)} is overly broadened, whereas names such \textit{ballads denomination, supermanager, thespian} might be overly specified to fit into table headers. \textit{(iii)} CTA incurs least semantic drift in RPL. In all baseline methods, there is a good chance to observe semantic-distinctive pairs such as \textit{(region, member)}, \textit{(type, number)}, \textit{(type, guy)}. With CTA, such risk is minimal.

\section{Experimental Details}

\subsection{Original Datasets statistics}

\begin{table}[t]
  \centering
  \scalebox{0.7}{
    \begin{tabular}{lcccccc}
    \toprule
    \multicolumn{1}{c}{\multirow{2}[4]{*}{Datasets}} & \multicolumn{3}{c}{Train} & \multicolumn{3}{c}{Dev} \\
\cmidrule(lr){2-4} \cmidrule{5-7}  & \textit{\#T} & \textit{\#Q} & \textit{\#Avg. Col}  & \textit{\#T} & \textit{\#Q}  & \textit{\#Avg. Col} \\
    \midrule
    WTQ   & $1,290$  & $9,030$ & $6.39$ & $327$  & $2,246$ & $6.41$ \\
    WikiSQL & $18,590$ & $56,355$ & $6.40$ & $2,716$  & $8,421$ & $6.31$ \\
    Spider & $795$   & $6,997$ & $5.52$ & $81$ & $1,034$ & $5.45$\\
    CoSQL & $795$  & $9,478$ & $5.52$ & $81$  & $1,299$ & $5.45$ \\
    SParC & $795$  & $12,011$ & $5.52$ & $81$ & $1,625$ & $5.45$ \\
    \bottomrule
    \end{tabular}%
    }
  \caption{Original datasets statistics. \textit{\#T} represents total number of tables in a dataset (\textit{\#Q} for questions). \textit{\#Avg. Col} stands for avg. number of columns per table. Spider, CoSQL and SParC share the same tables.}
  \label{tab:ori-statistics}%
\end{table}%

The detail statistics of five Text-to-SQL datasets are shown in Table \ref{tab:ori-statistics}.
According to CoSQL \citep{yu2019cosql} and SParC \citep{yu-etal-2019-sparc} paper, the two multi-turn Text-to-SQL datasets share the same tables with Spider \citep{yu-etal-2018-spider}.

\subsection{Baseline Details}
\label{sec:defense-impl-details}

\paragraph{SQLova}
For all defense results of the WikiSQL dataset, we employ the SQLova model, whose official codes are released in \citep{hwang2019comprehensive}.
We use uncased BERT-large as the encoder. 
The learning rate is $1 \times 10^{-3}$ and the learning rate of BERT-large is  $1 \times 10^{-5}$. 
The training epoch is 30 with a batch size of 12. 
The training process lasts 12 hours on a single 16GB Tesla V100 GPU.

\paragraph{SQUALL} 
We employ the SQUALL model, following \citep{shi-etal-2020-potential}, to get all defense results of the WTQ dataset.
The training epoch is 20 with a batch size of 30;
The dropout rate is 0.2;
The training process lasts 9 hours on a single 16GB Tesla V100 GPU.

\paragraph{ETA} 
We implement the ETA model following \citep{liu-etal-2021-awakening}.
We use an uncased BERT-large whole word masking version as the encoder. 
The learning rate is $5 \times 10^{-5}$ and the training epoch is 50. 
The batch size and gradient accumulation steps are 6 and 4.
The training process lasts 24 hours on a single 32GB Tesla V100 GPU.

\subsection{Attack Performance Calculation Details}
\label{sec:per-drop-detail}

\begin{table}[t!]
  \small
  \centering
    \begin{tabular}{ p{0.9cm}<{\centering} p{1cm}<{\centering} p{0.5cm}<{\centering} p{1.5cm}<{\centering} p{1.5cm}<{\centering}}
    \toprule
    \textbf{Dataset} & \textbf{Baseline} & \textbf{Dev}  &  \textbf{RPL}   & \textbf{ADD} \\
    \midrule
    \multirow{2}[2]{*}{Spider} & DuoRAT & $69.9$  & $23.8 \pm 2.1$ \tiny{\color{red}{(-46.1 / -65.9\%)}} & $36.4 \pm 1.3$ \tiny{\color{red}{(-33.5 / -47.9\%)}} \\
          & ETA   & $70.8$  & $27.6 \pm 1.8$ \tiny{\color{red}{(-43.2 / -61.0\%)}} & $39.9 \pm 0.9$ \tiny{\color{red}{(-30.9 / -43.6\%)}} \\
    
    \midrule
    \multirow{2}[2]{*}{WikiSQL} & SQLova & $81.6$  & $27.2 \pm 1.3$ \tiny{\color{red}{(-54.4 / -66.7\%)}} & $66.2 \pm 2.3$ \tiny{\color{red}{(-15.4 / -18.9\%)}}\\
        & CESQL & $84.3$  & $52.2 \pm 0.9$ \tiny{\color{red}{(-32.1 / -38.1\%)}}  & $71.2 \pm 1.5$ \tiny{\color{red}{(-13.1 / -15.5\%)}} \\
    \midrule 
    WTQ   & SQUALL & $44.1$  & $22.8 \pm 0.5$  \tiny{\color{red}{(-21.3 / -48.3\%)}} & $32.9 \pm 0.8$ \tiny{\color{red}{(-11.2 / -25.4\%)}} \\
    
    \midrule
    \multirow{2}[2]{*}{CoSQL} & EditSQL & $39.9$  & $13.3 \pm 0.7$ \tiny{\color{red}{(-26.6 / -66.7\%)}}  & $30.5 \pm 1.1$ \tiny{\color{red}{(-9.4 / -23.6\%)}} \\
          & IGSQL & $44.1$  & $16.4 \pm 1.2$ \tiny{\color{red}{(-27.7 / -62.8\%)}} & $32.8 \pm 2.1$ \tiny{\color{red}{(-11.3 / -25.6\%)}} \\
    \midrule
    \multirow{2}[2]{*}{SParC} & EditSQL & $47.2$  & $30.5 \pm 0.9$ \tiny{\color{red}{(-16.7 / -35.4\%)}}  & $40.2 \pm 1.2$ \tiny{\color{red}{(-7.0 / -14.8\%)}} \\
          & IGSQL & $50.7$  & $34.2 \pm 0.5$ \tiny{\color{red}{(-16.5 / -32.5\%)}} & $42.9 \pm 1.7$ \tiny{\color{red}{(-7.8 / -15.4\%)}} \\
    \bottomrule
    \end{tabular}%
  \caption{The Exact Match Accuracy on the development set and ADVETA. Red font denotes the absolute(left) and relative(right) performance drop percentage compared with original dev accuracy.
  }
  \label{tab:attack-results-with-relative-drop}%
\end{table}%

Table \ref{tab:attack-results-with-relative-drop} shows the attack performance of RPL and ADD perturbations.
In this section, we show the calculation details of the average attack relative performance drop.
For example, on the Spider dataset, the relative performance drop of the DuoRAT model against RPL perturbation is 65.9\%, and 61.0\% for the ETA model.
For RPL perturbation, we average out the relative performance drop of 9 models and report the average relative percentage drop ($53.1\%$). 
Same as RPL, we get the average relative percentage drop ($25.6\%$) for ADD perturbation.

\subsection{Schema Linking Calculation}
\label{sec:schema-linking-cal}
We follow the work of~\citet{liu-etal-2021-awakening} to measure the performance of ETA schema linking predictions.
Let $\Omega_{col}$ be a set $\{(c,q)_i|1\leq i\leq N\}$ which contains $N$ gold (column-question token) tuples.
Let $\overline{\Omega}_{col}$ be a set $\{(\overline{c},\overline{q})_j|1\leq j\leq M\}$ which contains $M$ predicted (column-question token) tuples.
We define the precision(\colprec), recall(\colrecall), F1-score(\colfscore) as:
\[
\frac{\left|\Gamma_{col}\right|}{\left|\overline{\Omega}_{col}\right|}, \frac{\left|\Gamma_{col}\right|}{\left|{\Omega}_{col}\right|}, \frac{2\text{Col}_P \text{Col}_R}{\text{Col}_P + \text{Col}_R}
\]
where ${\Gamma}_{col} = {\Omega}_{col} \bigcap {\overline{\Omega}_{col}}$.
The definitions of \tabprec, \tabrecall, \tabfscore are similar.

\section{Baseline Approach Details}
\label{a:baseline_details}

\paragraph{W2V} To generate candidates for a given column, W2V randomly samples $five$ candidates from the top 15 cosine-similar (Numberbatch word embeddings) for RPL and 15-50 for ADD. Textfooler and BERT-Attack also follow this hyper-parameter setting. For both TextFooler and BERT-Attack, we skip their word importance ranking (WIR) modules while only keeping the word transformer modules for candidate generation\footnote{We contextualize columns with templates that additionally considers cell values and POS-tag consistency.}.

\paragraph{TextFooler} TextFooler is one of the state-of-the-art attacking frameworks for discriminative tasks on unstructured text. 
We skip its word importance ranking (WIR) step since our target column has already been located. 
Its word transformer module is faithfully re-implemented to generate candidates for a target column. 
Counter-fitted word embedding \cite{DBLP:journals/corr/MrksicSTGRSVWY16} are used for similarity computation, and modified sentences are constrained by both POS-tag consistency and Sim-CSE \cite{DBLP:journals/corr/abs-2104-08821} similarity score. 

\paragraph{BERT-Attack} BERT-Attack is another representative text attacking framework.
Similar to our adaptation of TextFooler, we skip WIR and only keep the core masked language model-based word transformation. 
Following original work, low-quality or sub-word tokens predicted by BERT-Large are discarded; perturbed sentence similarities compared with the original are guaranteed by Sim-CSE.